\pdfoutput=1
\documentclass[11pt]{article}

\usepackage[preprint]{acl}

\PassOptionsToPackage{table}{xcolor}
\usepackage{times}
\usepackage{latexsym}
\usepackage{booktabs}
\usepackage{amsmath}
\usepackage{caption}
\usepackage{subcaption}
\usepackage[frozencache,cachedir=.]{minted}
\usepackage{listings}
\usepackage{xcolor}

\usepackage{xurl}

\newcommand{\TMM}{\texttt{Text}_{\text{MM}}}
\newcommand{\VMM}{\texttt{Vis}_{\text{MM}}}

\usepackage{cuted}
\usepackage{hyperref}
\newcommand{\adjustimg}{
  \hspace*{\dimexpr\evensidemargin-\oddsidemargin}
}
\newcommand{\centerimg}[2][width=\textwidth]{
  \makebox[\textwidth]{\adjustimg\includegraphics[#1]{#2}}%
}

\usepackage[T1]{fontenc}

\usepackage[utf8]{inputenc}

\usepackage{microtype}

\usepackage{inconsolata}

\usepackage{graphicx}
\usepackage{subcaption}
\usepackage{tabularray}
\usepackage{multirow}
\usepackage{array}
\usepackage{dsfont}
\usepackage{xcolor}
\usepackage{tikz} 
\usepackage{booktabs}

\title{Beyond Pattern Recognition: Probing Mental Representations of LMs}

\author{
    Moritz Miller  \quad Kumar Shridhar \\
    \normalsize 
    \{\href{mailto:millerm@ethz.ch}{\texttt{millerm}}, \href{mailto:shkumar@ethz.ch}{\texttt{shkumar}}\}@ethz.ch \\
    \vspace{10pt} 
    \fbox{\includegraphics[width=.15\linewidth]{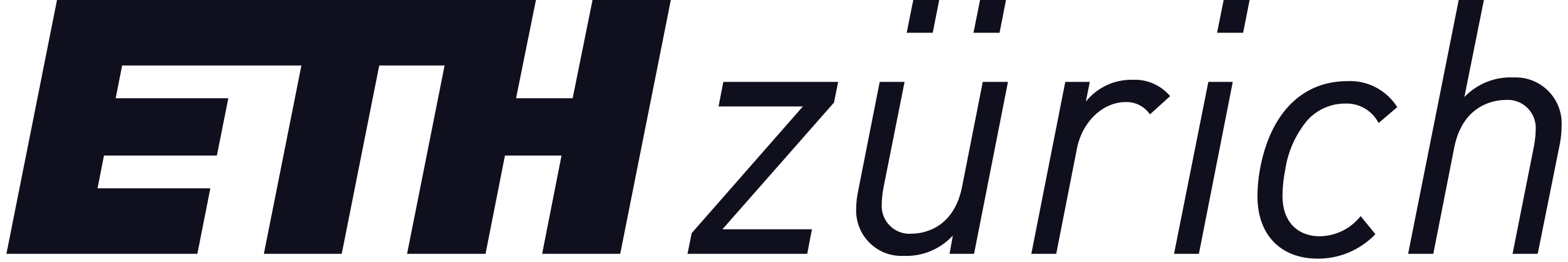}}
}

\begin{document}
\maketitle

\begin{strip}
  \vspace{-58pt}

    \noindent\centerimg[width=\linewidth]{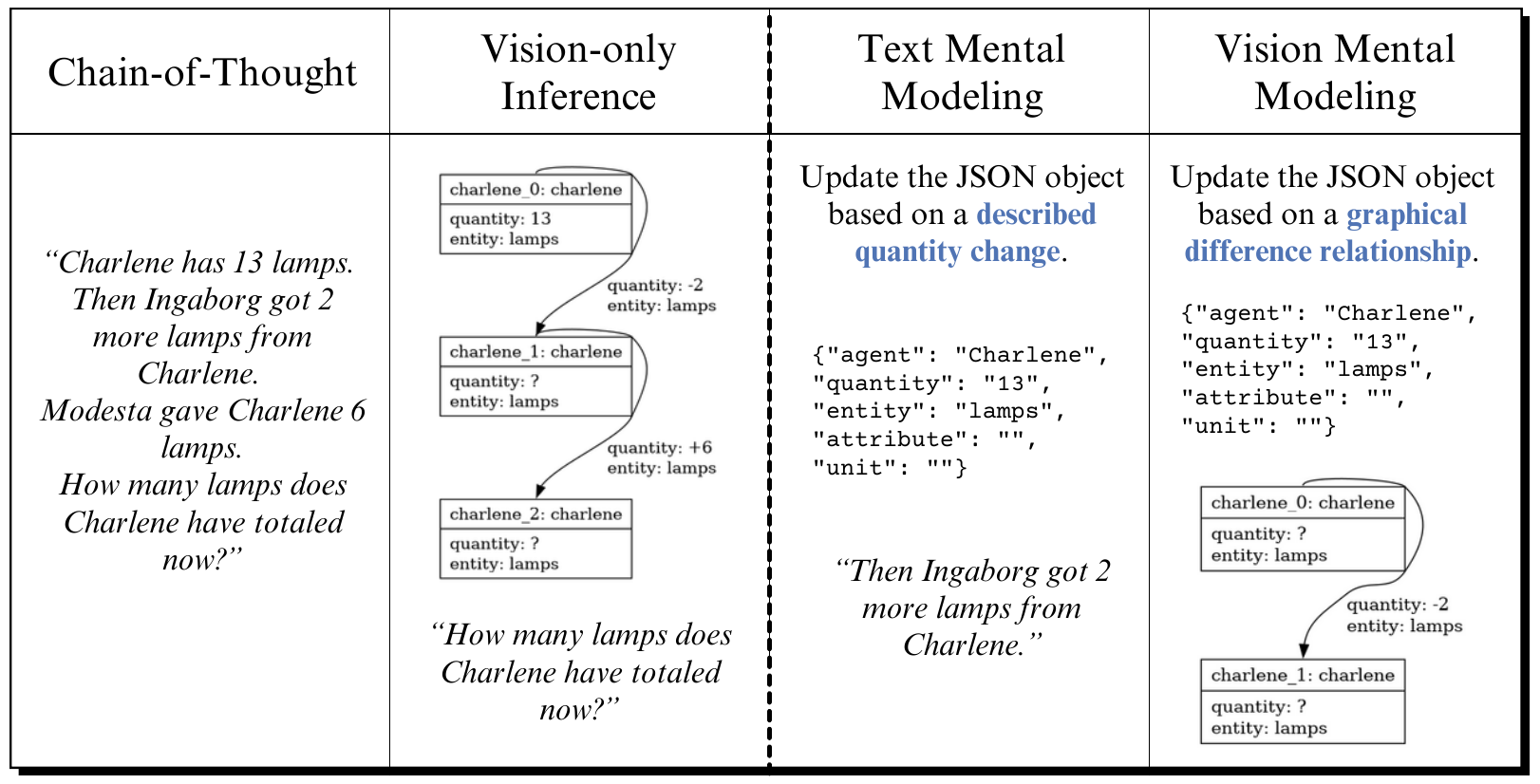}

  \fontsize{10pt}{12pt}\selectfont
    Figure 1: The mental modeling approach enables a Language Model to generate accurate answers through creating intermediate representations of the provided information. All four instances represent the inputs to the model.
  \label{fig:concept}

\end{strip}

\setcounter{figure}{1}

\begin{abstract}

Language Models (LMs) have demonstrated impressive capabilities in solving complex reasoning tasks, particularly when prompted to generate intermediate explanations. However, it remains an open question whether these intermediate reasoning traces represent a dynamic, evolving thought process or merely reflect sophisticated pattern recognition acquired during large-scale pre-training. Drawing inspiration from human cognition—where reasoning unfolds incrementally as new information is assimilated and internal models are continuously updated—we propose to delve deeper into the mental model of various LMs. We propose \emph{a new way to assess the mental modeling} of LMs, where they are provided with problem details gradually, allowing each new piece of data to build upon and refine the model’s internal representation of the task. We systematically compare this step-by-step mental modeling strategy with traditional full-prompt methods across both text-only and vision-and-text modalities. Experiments on the \textsc{MathWorld} dataset across different model sizes and problem complexities confirm that both text-based LLMs and multimodal LMs struggle to create mental representations, questioning how their internal cognitive processes work. Code and data are available at \url{https://github.com/moXmiller/mental-modeling.git}.

\end{abstract}

\section{Introduction}
Language Models have demonstrated remarkable capabilities in tackling complex reasoning tasks, particularly when they are prompted to generate intermediate explanations~\cite{cot,letsthinksbs,tocotornot}. By decomposing problems into smaller steps, intermediate reasoning has led to significant performance improvements across a range of domains, from arithmetic to commonsense reasoning~\cite{subques,cot}. Despite empirical successes, a fundamental question remains: do these intermediate reasoning traces reflect a genuine, evolving mental process, or are they simply the result of sophisticated pattern recognition learned from large-scale training data?

In human cognition, a key principle of reasoning is that it unfolds dynamically. As we read or process new information, we continuously build and update an internal ``mental model'' of the situation at hand~\cite{siegler,johnson2013mental}. While mental modeling is a broad concept that has been discussed under various frameworks, the central idea is that humans integrate new details into an evolving representation of the problem, revising earlier assumptions as needed to form a coherent understanding~\cite{johnson2013mental}. We formulate a Cognitive Science hypothesis: \emph{If LMs can truly simulate human-like incremental reasoning, then providing them with information in a step-by-step manner should lead to iterative refinements that mirror how humans dynamically update their mental models as they read and process text.} Testing this hypothesis requires evaluating whether an LM’s intermediate reasoning steps evolve in a coherent and logically consistent way as new details are introduced.
    
To rigorously evaluate this approach, we conduct experiments on \textsc{MathWorld}~\cite{opedal} (and its extended version). This dataset is specifically curated to assess both memory and reasoning abilities under varying levels of problem complexity and across different model scales. By comparing the performance of various LMs using our step-by-step mental modeling method against traditional full-prompt strategies, we aim to uncover whether LMs are truly capable of simulating human-like reasoning. Crucially, our study goes beyond merely examining final answers. We also analyze the quality and coherence of the models’ intermediate reasoning steps to gain deeper insights into their underlying cognitive processes.

\section{Related Work}

\texttt{CoT} prompting has emerged as a transformative technique in the domain of large language models (LLMs). Wei et al.~\cite{cot} demonstrated that eliciting intermediate reasoning steps significantly improves performance on complex tasks such as arithmetic and commonsense reasoning. However, subsequent work has raised important questions about whether these generated reasoning traces reflect a genuine internal problem-solving process or are simply a by-product of pattern recognition over extensive pre-training data~\cite{turpin2023,epflfaithfulness}.

A related line of inquiry investigates the internal mechanisms underlying reasoning in both text-only and multimodal settings. Early research in vision-language understanding established that aligning visual and textual representations is key to coherent reasoning. For instance, models such as LXMERT~\cite{tan-bansal-2019-lxmert}, VilBERT~\cite{lu2019vilbert}, and UNITER~\cite{chen2020uniter} leveraged cross-modal attention mechanisms to learn shared representations, thereby enhancing performance on tasks that require integrating visual cues with textual context. Similarly, contrastive learning approaches, as exemplified by CLIP~\cite{radford2021learning}, have effectively mapped images and text into a common embedding space, underscoring the importance of alignment in developing a unified mental model.

Mental modeling in LLMs—forming internal task or world representations—has been explored in both text-only and multimodal contexts. In text-only settings, models must infer structure solely from textual descriptions, often lacking coherent representations of objects and their relations~\cite{gu2023}. In contrast, multimodal models can leverage visual grounding to enhance internal representations~\cite{gao2024}. Other strategies facilitate reasoning through iterative dialogue between image and language models~\cite{zeng2022socratic} or by translating visual information into natural language~\cite{lu22, mmcot}. Recent advances in multimodal reasoning extend beyond simple image-to-text transcriptions, focusing instead on reasoning across multiple modalities simultaneously~\cite{mvot}. 
These findings align with classic cognitive insights~\cite{larkin1987diagram} suggesting that visual representations reduce cognitive load by making implicit relationships explicit. Collectively, this body of research underscores that multimodal models construct richer and more grounded mental models than their text-only counterparts.

The evolution toward large-scale multimodal models—such as Flamingo~\cite{alayrac2022flamingo} and BLIP-2~\cite{li2023blip}—has further underscored that even when visual inputs are available, the core reasoning engine remains largely text-based. Proprietary systems like GPT-4V~\cite{openai2024gpt4} and PaLM-E~\cite{driess2023palm} reinforce this observation: despite their ability to process visual information, these models often exhibit reasoning processes that are strongly influenced by their language model components. This body of work suggests that whether operating on text alone or in combination with visual modalities, LLMs tend to develop a unified internal representation that drives their reasoning capabilities.

Finally, some recent research has explored how sequential or incremental information delivery influences a model’s reasoning process. Iterative refinement techniques~\cite{nye2021show} and self-consistency methods~\cite{wang2023self} indicate that presenting information step-by-step allows models to update their internal state and progressively build a more accurate understanding of a problem. These findings resonate with cognitive theories of human reasoning, where information is assimilated gradually to form and refine a mental model~\cite{siegler}. 

Our work builds upon these diverse research strands by introducing a mental model approach that explicitly supplies information incrementally. By comparing this approach to traditional full-prompt \texttt{CoT} methods on datasets such as \textsc{MathWorld}, we aim to investigate whether LLMs can truly evolve their internal representations in a human-like, dynamic fashion—or whether their reasoning remains predominantly a product of pattern recognition. In doing so, we not only extend prior studies on \texttt{CoT} and multimodal alignment but also probe the fundamental nature of reasoning in LLMs.

\section{Methodology}
\label{sec:mental_model}
Let's say we have a reasoning dataset $\mathcal{D} = \{(X_i, Y_i, d_i)\}_{i=1}^N$ consisting of $N$ problems, where
$X_i$ is the \emph{complete} text of the $i$-th problem, $Y_i$ is the ground-truth final solution for problem $i$, and $d_i$ is the number of incremental steps needed to solve the problem $i$ (denoted as depth in our work which ranges from 1 to 6).
Including the initial sentence, each problem $X_i$ can be split into $d_i + 1$ increments: $X_i \;=\; \bigl\langle\, X_i^{(0)} \oplus X_i^{(1)} \oplus X_i^{(2)} \oplus \cdots \oplus X_i^{(d_i)} \bigr\rangle,
$ where $\oplus$ denotes string (or token) concatenation.

A language model (LM) with parameters $\theta$ is denoted by $f_\theta(P) \rightarrow O$,
which produces an output (token sequence) $O$ in response to a prompt $P$.
\paragraph{Baseline: Chain-of-Thought}
For an approach like Chain-of-Thought (\texttt{CoT}), each problem is given to the model \emph{all at once}. That is, we format the entire text $X_i$ of problem $i$ (including all $d_i + 1$ steps) into a single prompt $
P_i^\text{full} \;=\; \texttt{FormatFull}(X_i)$,
where \texttt{FormatFull} also include instructions to solve the task. The full set of instructions is provided in Appendix \ref{app:prompts}. Then, the LM generates a chain of thought plus a final answer:
\[
    O_i^\text{full} \;=\; f_\theta\bigl(P_i^\text{full}\bigr)
\]
We parse $O_i^\text{full}$ using regex to extract the final predicted answer $ \hat{Y}_i^\text{full} \;=\; \mathrm{parse}\bigl(O_i^\text{full}\bigr)$ and calculate the accuracy over it. 

\paragraph{Mental Modeling}
By providing the input prompt incrementally to the LM, we obtain the output $O$ at each step. For example, at step 0, we only provide the initial information as the prompt,
\[
        P_{i,0}^\text{mm} \;=\; \texttt{FormatStep}\bigl(X_i^{(0)}\bigr).
\]
 The corresponding output contains the starting mental model representation for Step 0,
 \[
        O_{i,0}^\text{mm} \;=\; f_\theta\bigl(P_{i,0}^\text{mm}\bigr).
\]
Similarly, for all steps $k$ in $1 \le k \le d_i$, we only provide \emph{the last output or updated mental model} plus the next piece of information,
    \[
        P_{i,k}^\text{mm} \;=\; \texttt{FormatStep}\Bigl(X_i^{(k)}, \; O_{i,k-1}^\text{mm}\Bigr).
    \]
such that the model updates its mental representation output,
    \[
        O_{i,k}^\text{mm} \;=\; f_\theta\bigl(P_{i,k}^\text{mm}\bigr).
    \]
Note that $O_{i,k-1}^\text{mm}$ is updated incrementally to comprise all relevant past reasoning. That way, only the last updated mental model representation is provided at each step. For multimodal LMs (text + vision), we provide the last output $O$ in image format while the question input $X$ is in textual form. An example is demonstrated in Appendix \ref{app:prompts} for both text only LMs and multimodal LMs. 

Finally, the $d_i$-th output, $O_{i,d_i}^\text{mm}$ is parsed using regex to extract the final predicted answer,
\begin{align*}
    \hat{Y}_i^\text{mm} \;=\; \mathrm{parse}\Bigl(O_{i,d_i}^\text{mm}\Bigr).
\end{align*}

\newcommand{\colscore}[2]{%
  \pgfmathsetmacro{\gb}{max(0,0.6-0.9*abs(#2))}%
  \textcolor[rgb]{1,\gb,\gb}{#1}%
}

\begin{table*}[ht]
\centering
\resizebox{\linewidth}{!}{%
\begin{tabular}{c l l c c c c c c}
    \toprule
    & \textbf{Model} & \textbf{Strategy} & \textbf{Depth 1} & \textbf{Depth 2} & \textbf{Depth 3} & \textbf{Depth 4} & \textbf{Depth 5} & \textbf{Depth 6} \\
    \midrule
    \multirow{5}{*}{\rotatebox{90}{Text only}} 
      & \multirow{2}{*}{Llama3.2-3B}
          & Chain-of-Thought (\texttt{CoT}) 
              & 0.94 & 0.93 & 0.91 & 0.81 & 0.70 & 0.64 \\[3mm]
      & 
          & Text Mental Modeling ($\TMM$) 
              & \colscore{0.48}{0.46}
              & \colscore{0.23}{0.70}
              & \colscore{0.13}{0.78}
              & \colscore{0.08}{0.73}
              & \colscore{0.07}{0.63}
              & \colscore{0.05}{0.59} \\
    \cmidrule(lr){2-9}
      & \multirow{2}{*}{Llama3.3-70B}
          & Chain-of-Thought (\texttt{CoT}) 
              & 0.96 & 0.95 & 0.97 & 0.97 & 0.93 & 0.91 \\[3mm]
      & 
          & Text Mental Modeling ($\TMM$) 
              & \colscore{0.75}{0.21}
              & \colscore{0.59}{0.36}
              & \colscore{0.46}{0.51}
              & \colscore{0.42}{0.55}
              & \colscore{0.30}{0.63}
              & \colscore{0.27}{0.64} \\
    \midrule
    \multirow{5}{*}{\rotatebox{90}{Distilled}} 
      & \multirow{2}{3cm}{DeepSeek-R1-Distill-Qwen-32B}
          & Chain-of-Thought (\texttt{CoT}) 
              & 0.98   & 0.99   & 1.00   & 0.99   & 0.99   & 0.99 \\[3mm]
      & 
          & Text Mental Modeling ($\TMM$) 
              & \colscore{0.90}{0.08}
              & \colscore{0.83}{0.16}
              & \colscore{0.73}{0.27}
              & \colscore{0.67}{0.32}
              & \colscore{0.64}{0.35}
              & \colscore{0.62}{0.37} \\
    \cmidrule(lr){2-9}
      & \multirow{2}{3cm}{DeepSeek-R1-Distill-Llama-70B}
          & Chain-of-Thought (\texttt{CoT}) 
              & 0.92   & 0.95   & 0.98   & 0.99   & 1.00   & 1.00 \\[3mm]
      & 
          & Text Mental Modeling ($\TMM$) 
              & 0.99
              & 0.99
              & 0.98
              & \colscore{0.97}{0.02}
              & \colscore{0.98}{0.02}
              & \colscore{0.97}{0.03} \\
    \midrule
    \multirow{7}{*}{\rotatebox{90}{Multimodal}} 
      & \multirow{3}{*}{Llama3.2-11B}
          & Chain-of-Thought (\texttt{CoT}) 
              & 0.90 & 0.96 & 0.96 & 0.96 & 0.95 & 0.94 \\[3mm]
      & 
          & Text Mental Modeling ($\TMM$) 
              & \colscore{0.22}{0.68}
              & \colscore{0.10}{0.86}
              & \colscore{0.06}{0.90}
              & \colscore{0.03}{0.93}
              & \colscore{0.02}{0.93}
              & \colscore{0.02}{0.92} \\[1mm]
    & 
          & Vision Mental Modeling ($\VMM$) 
              & 0.95
              & \colscore{0.92}{0.04}
              & \colscore{0.86}{0.10}
              & \colscore{0.84}{0.12}
              & \colscore{0.84}{0.11}
              & \colscore{0.79}{0.15} \\
    \cmidrule(lr){2-9}
      & \multirow{3}{*}{Llama3.2-90B} 
          & Chain-of-Thought (\texttt{CoT}) 
              & 0.99 & 0.99 & 0.99 & 0.99 & 0.99 & 0.99 \\[3mm]
      & 
          & Text Mental Modeling ($\TMM$) 
              & \colscore{0.72}{0.27}
              & \colscore{0.57}{0.42}
              & \colscore{0.42}{0.57}
              & \colscore{0.32}{0.67}
              & \colscore{0.26}{0.73}
              & \colscore{0.20}{0.79} \\[1mm]
      & 
          & Vision Mental Modeling ($\VMM$) 
              & \colscore{0.85}{0.14}
              & \colscore{0.73}{0.26}
              & \colscore{0.66}{0.33}
              & \colscore{0.63}{0.36}
              & \colscore{0.48}{0.51}
              & \colscore{0.45}{0.54} \\
    \bottomrule
\end{tabular}%
}
\caption{Accuracy comparison across reasoning depths for Chain-of-Thought (\texttt{CoT}), Text Mental Modeling ($\TMM$), and Vision Mental Modeling ($\VMM$) across different models. For alternative strategies, scores are color-coded based on their difference from the \texttt{CoT} baseline. Larger deviations are shown in darker shades of red.}
\label{tab:acc}
\end{table*}

\subsection{Experimentation Details}
\label{subsec:dataset}
\paragraph{\textsc{MathWorld} Dataset}  
We utilize the \textsc{MathWorld} dataset~\cite{opedal}, which consists of a diverse collection of mathematical story problems. Its flexibility allows for the dynamic generation of math problems and the creation of graphical representations of the underlying content. This enables us to assess mental modeling capabilities in both natural language and visual contexts.  
For this study, we focus on linear "transfer" problems, where each sentence represents a zero-sum exchange between two individuals. We generated 400 problems per depth level, resulting in a total of 2,400 examples for our experiments.

\paragraph{Evaluation Metrics}
We calculate the final accuracy $\mathrm{Acc}$ defined across the entire dataset $\mathcal{D}=\{(X_i,Y_i)\}_{i=1}^N$ as:
\[
    \mathrm{Acc} \;=\; \frac{1}{N} \sum_{i=1}^{N} \mathds{1}\!\bigl(\hat{Y}_i \;=\; Y_i\bigr),
\]
where $\hat{Y}_i$ be the predicted final answer for problem $i$,  $Y_i$ be the ground-truth answer, and $\mathds{1}(\cdot)$ is the indicator function, which equals 1 if the predicted answer equals the ground truth, and 0 otherwise.

\paragraph{Model Selection}
\label{subsec:models}
To assess the mental modeling capacity, we compare the performance of text only LMs with text-and-vision multimodal models. Our text models include Llama models \cite{dubey2024llama} of sizes 3B and 70B while multimodal versions include Llama models of sizes 11B and 90B. We also used the distilled version of Qwen 32B and Llama 70B from DeepSeek R1 \cite{guo2025deepseek} as released by them. In Appendix \ref{app:setup}, we discuss the specific model configurations.

\begin{figure}
    \centering
    \includegraphics[width=\linewidth]{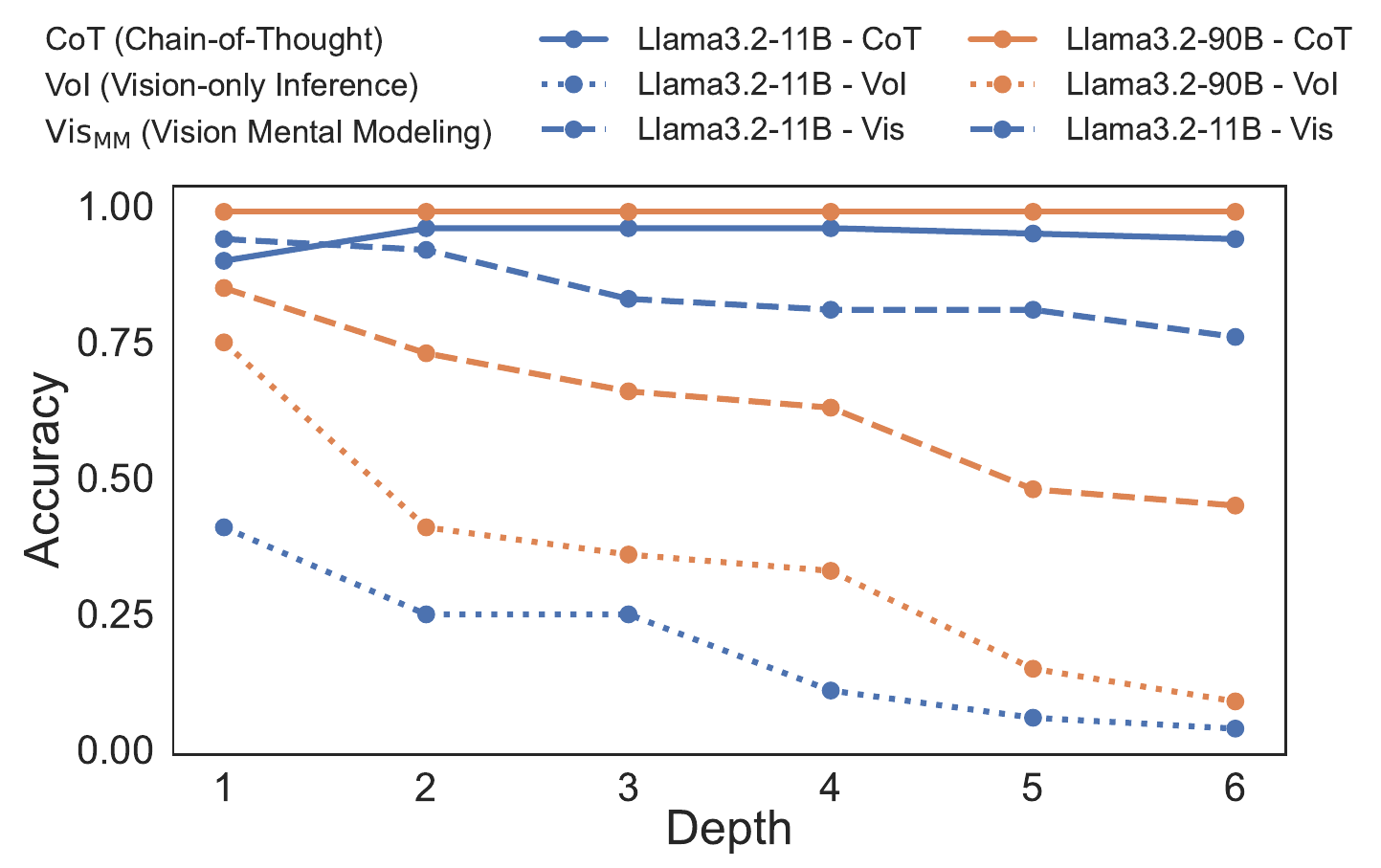}
    \caption{Accuracy of \texttt{CoT}, \texttt{VoI}, and $\VMM$ over increasing depth across models Llama3.2-11B and Llama3.2-90B. Line styles differentiate prompting techniques, while colors distinguish model sizes.}
    \label{fig:VMM_VoI_vot}
\end{figure}

\section{Results and Discussion}
\paragraph{LLMs excel at pattern recognition, not mental modeling} 

\autoref{tab:acc} presents the final accuracy across problems of increasing complexity, up to depth 6. As anticipated, when models are given the complete problem statement and prompted using \texttt{CoT} reasoning, they achieve near-perfect accuracy across all difficulty levels. Only at Llama 3B does accuracy decline from 0.94 at depth 1 to 0.64 at depth 6. This is likely due to its limited reasoning capacity. However, when required to incrementally construct and refine a mental model, all models--except for the distilled Llama70B--exhibit a notable drop in accuracy. Too, most models show a consistent decline in performance as problem depth increases. Only distilled Llama 70B maintains stable accuracy across varying complexity. This suggests that model distillation from a stronger source enhances the ability to perform mental modeling, a capability that is otherwise challenging to develop.

\paragraph{Vision Improves Mental Modeling}

For multimodal models, incorporating a vision-based mental model, where the input is represented as an image and updated incrementally with new information, proves highly beneficial. As shown in \autoref{tab:acc}, $\VMM$ consistently outperforms $\TMM$ for both Llama 11B and 90B. Notably, Llama 11B demonstrates significantly better performance with $\VMM$ compared to Llama 90B. Manual inspection suggests that Llama 90B overfits the visual task, often hallucinating an image rather than utilizing the provided one. Details are given in Appendix \ref{app:overfitting}.

\paragraph{Model-Based Mental Modeling Differs from Human Cognition}  

In \autoref{fig:VMM_VoI_vot}, we present all intermediate steps as visual representations of a mental model and task multimodal models with solving the final step. This approach, referred to as Vision-only Inference (\texttt{VoI}), performs significantly worse than \texttt{CoT} reasoning. As mentioned, larger models underperform compared to smaller ones, likely due to overfitting. However, Vision Mental Modeling ($\VMM$) outperforms \texttt{VoI}, indicating that the way models construct mental representations differs from human visualization-based reasoning. Additional examples illustrating these differences are provided in Appendix \ref{app:prompts}.

\paragraph{Modality Switching in Multimodal Models}

\begin{figure}
    \centering
    \includegraphics[width=\linewidth]{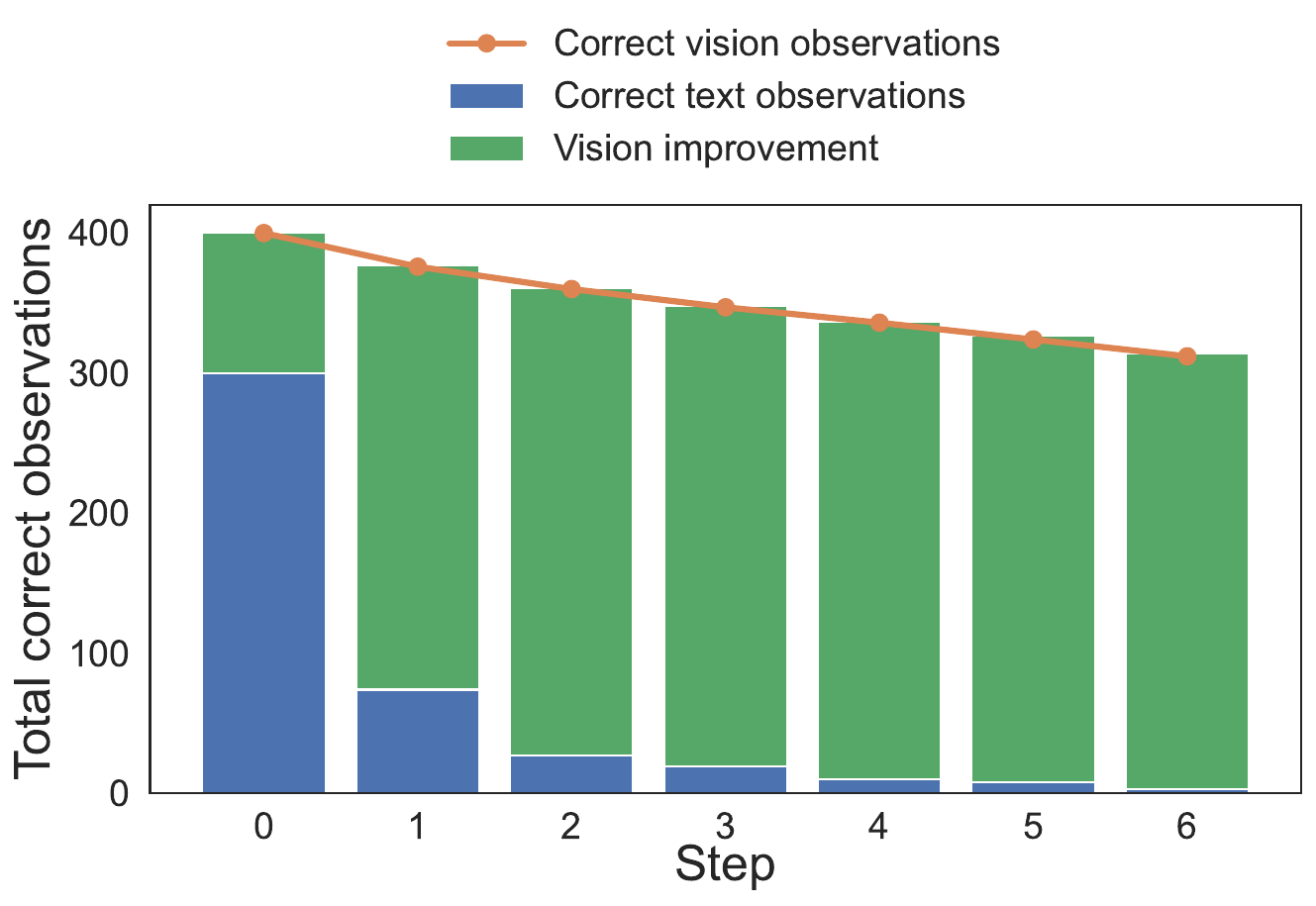}
    \caption{Correct predictions at each intermediate step for depth 6 problems in Llama3.2-11B. Blue represents $\TMM$, while green indicates $\VMM$ applied to all incorrect predictions as identified by an Oracle verifier.
    }
    \label{fig:alignment}
\end{figure}

Ideally, a multimodal model should leverage both textual and visual modalities, dynamically selecting the most effective one based on the given problem. To investigate this, we implement an Oracle verifier that identifies instances where the text modality ($\TMM$) produces an incorrect answer and switches to the vision modality ($\VMM$) instead.
\autoref{fig:alignment} illustrates cases where $\TMM$ errors trigger a modality switch to $\VMM$. Interestingly, the final performance remains identical to using $\VMM$ alone (depicted by the green bar aligning with the orange line), suggesting that whenever $\TMM$ was correct, $\VMM$ was also correct. This finding raises important questions about the alignment of mental representations between $\TMM$ and $\VMM$—specifically, whether improving this alignment could enhance $\TMM$’s performance. We leave this exploration for future work.

\section{Conclusion}

Our study on step-by-step \textit{mental modeling} reveals key limitations in current language models:

\begin{enumerate}
    \item While LMs perform well on single-prompt \texttt{CoT} reasoning, they struggle to develop and maintain dynamic internal representations when information is introduced incrementally.
    \item Vision-based mental modeling ($\VMM$) consistently outperforms text-based mental modeling ($\TMM$), indicating that incorporating visual components could significantly enhance reasoning capabilities in future models.
    \item In some cases, smaller multimodal models outperform their larger counterparts, suggesting that larger models may suffer from overfitting or hallucination.
    \item The way models construct mental representations differs from human cognition, implying that alternative strategies for encoding and processing incremental context may be necessary rather than relying on human-like reasoning structures.
\end{enumerate}

\section*{Limitations}

Using the \textsc{MathWorld} framework, our study is inherently constrained by the limitations imposed by the dataset. In this context, we focus exclusively on "transfer" problems, which represent only a subset of the available data. However, given the model's difficulties with this relatively simple framework, our choice of dataset remains justified. Furthermore, our approach to measuring mental modeling relies on the model's ability to construct and update JSON objects. After exploring various strategies for assessing mental modeling, this method has yielded the most reliable results. Nevertheless, it remains plausible that the model employs an alternative form of mental representation that is not readily interpretable.

\section*{Ethical Considerations}

In our judgment, this study does not raise substantial ethical concerns. Primarily, our focus is on analyzing the mental modeling capabilities of language models. Since this is largely a descriptive task, we ensure that no inappropriate or violent math story problems are generated in the process.

\section*{Acknowledgments}

We would like to express our gratitude to Andreas Opedal and Haruki Shirakami for granting us access to the \textsc{MathWorld} dataset and assisting in generating the examples and setting up initial experiments. We also appreciate Mrinmaya Sachan for his valuable feedback.

\bibliography{custom}

\appendix

\begin{figure}
    \centering
     \begin{subfigure}[b]{0.23\textwidth}
         \centering
         \includegraphics[width=0.8\linewidth]{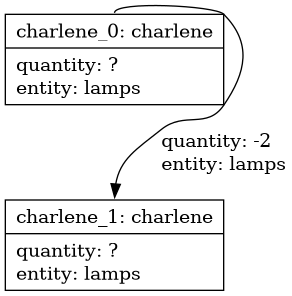}
    
         \caption{Step 1}
         \label{fig:step1}
     \end{subfigure}
    \hfill
     \begin{subfigure}[b]{0.23\textwidth}
         \centering
         \includegraphics[width=0.8\linewidth]{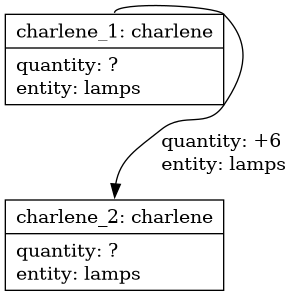}
    
         \caption{Step 2}
         \label{fig:step2}
     \end{subfigure}
     \caption{Graphical representation of difference relationships for a depth 2 problem in Vision Mental Modeling.}
    \label{fig:steps}
\end{figure}

\section{Prompt Details}
\label{app:prompts}
 As illustrated in \hyperref[fig:concept]{Figure 1}, we sequentially prompt the models to update their mental representations. Specifically, we begin by providing the first sentence, which contains the initial quantity possessed by the agent. Based on this input, the model generates a JSON object with the keys \texttt{"agent"}, \texttt{"quantity"}, \texttt{"entity"}, \texttt{"attribute"}, and \texttt{"unit"}. This JSON representation is then extracted and fed back into the model. For $\TMM$, the subsequent sentence is provided, whereas for $\VMM$, the next image is supplied (see \autoref{fig:steps}). Notably, the initial task remains identical for both text and vision modalities, with divergence occurring only when describing differences in quantity. 

Listing \ref{fig:prompt-VMM} presents a comprehensive $\VMM$ layout, including the problem statement, various prompt types, and the final output. This figure showcases a scenario in which the model successfully arrives at the correct solution. Moreover, Listings \ref{fig:prompt-TMM} and \ref{lst:python} summarize the output of $\TMM$. For improved readability, we have truncated the initial output (see \ref{fig:prompt-TMM}) and provided it separately (\ref{lst:python}). Here, python code is returned to contruct the JSON object. We also note that the model misunderstands the sentence: \texttt{Then Ingaborg got 2 more lamps from Charlene}. Interestingly, this is consistent with its \texttt{CoT} output (see \ref{fig:prompt-cot}). Furthermore, Listing \ref{fig:prompt-VoI} displays the \texttt{VoI} output for models with 11B and 90B parameters. In the case of Llama3.2-11B, the generation process begins with an incorrect answer, followed by iterative reasoning until the initial answer aligns with the derived solution. Although the model reaches the correct solution after five steps, it continues generating additional responses, appending terms such as "final final," "final final final," and "final final final final." Lastly, Listing \ref{fig:prompt-cot} illustrates the traditional \texttt{CoT} output. This example highlights a failure case for the smaller model, which, despite achieving near-perfect accuracy across varying complexities, struggles with sentences in which the relevant agent is mentioned last and their quantity is reduced.

\begin{figure}
    \centering
    \includegraphics[width=0.44\linewidth]{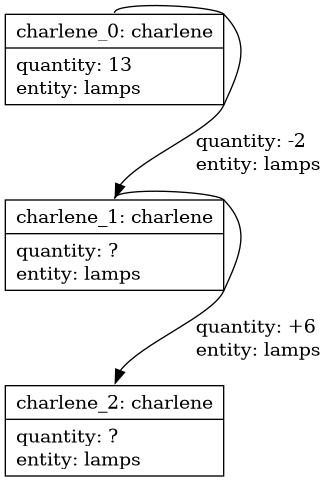}
    \caption{Complete graph representation for Vision-only Inference.}
    \label{fig:completegraph}
\end{figure}

\section{Model Configuration and Additional Experiments}
\label{app:setup}

\subsection{Setup and Model Configurations}
We run all experiments using vLLM and set the sampling parameters \texttt{temperature} $=$ $0$ and \texttt{output\_length} $=$ $2048$. In doing so, we employ Copilot inside Visual Studio Code. Additionally, we employ four Nvidia A100-80GB GPUs for inference over the multimodal models and distilled versions, and one Nvidia RTX3090 for the smaller models up to 8B parameters.

\subsection{Experiments on Other Models}
To ensure robust results, we conducted experiments across the entire suite of Llama3.2 Instruct models. For the 1B version, we observed that the model successfully extracts the JSON representation from the first sentence in only 21.75\% of cases. However, at the subsequent step, the model fails to update its mental representation in every case.

Furthermore, the Llama3.1 Instruct models tend to default to generating Python code for JSON extraction using regular expressions. For example, Llama3.1-8B frequently enters a loop of code generation, losing sight of the initial prompt. Compare also Listings \ref{fig:prompt-TMM} and \ref{lst:python} for a similar behavior in Llama3.2-11B.

Therefore, we restrict our analysis to Llama3.2 and Llama3.3 models of at least 3B parameters.

\begin{listing}[!ht]
\begin{minted}[fontsize=\footnotesize, frame=single, breaklines]{text}
This is the image: 
https://i.stack.imgur.com/6Qq0q.png
**Step 1: Extract the full JSON object.**
The JSON object is already provided: 
"{"agent": "Rosalind", "quantity": 8,
"entity": "spoons", "attribute": "",
"unit": ""}".

**Step 2: Infer the future quantity
using the information from the image.**
The image shows a decrease of 2 spoons. 
Therefore, the future quantity will be
the current quantity minus 2.

**Step 3: Return the updated JSON object.**
The updated JSON object with 
the future quantity is: 
"{"agent": "Rosalind", "quantity": 6,
"entity": "spoons", "attribute": "", 
"unit": ""}".

**Answer:** "{"agent": "Rosalind", 
"quantity": 6, "entity": "spoons", 
"attribute": "", "unit": ""}"
\end{minted}
\caption{Hallucination of Llama3.2-90B when instructed to solve intermediate $\VMM$ steps.}
\label{lst:hallucination}
\end{listing}

\section{Overfitting in Llama3.2-90B}
\label{app:overfitting}
In the $\VMM$ setting, Llama3.2-90B appears unable to fully comprehend the task. Despite being provided with the image, the model consistently generates the text \texttt{"This is the image:"}, followed by a URL that remains identical across different prompts. On average, this behavior occurs 116 times per 400-observation sample across all depths and strictly positive steps. For reference, see Listing \ref{lst:hallucination}. Notably, despite this repetitive output pattern, the model still correctly returns the updated JSON object.

\begin{listing*}

\begin{small}
\captionsetup{justification=centering, labelfont=bf, font=small}
\begin{minted}[fontsize=\footnotesize, frame=lines, framesep=2mm, baselinestretch=1.2, breaklines, breaksymbolleft={}, breaksymbolright={}]{text}
Problem: Charlene has 13 lamps. Then Ingaborg got 2 more lamps from Charlene. Modesta gave Charlene 6 lamps. How many lamps does Charlene have totaled now?

Initial Prompt: <|image|><|begin_of_text|> Extract structured information from the following sentence and update an empty JSON object with the categories:
'agent', 'quantity', 'entity', 'attribute', and 'unit'. If a category is not applicable, leave it as an empty string. Do not introduce additional categories. This is the sentence: 'Charlene has 13 lamps.'.

Initial Output: The information in the sentence is structured as follows:
* Agent: Charlene
* Quantity: 13
* Entity: lamps
* Attribute: None
* Unit: None
The structured information is represented as a JSON object:
{"agent": "Charlene", "quantity": 13, "entity": "lamps", "attribute": "", "unit": ""}

Update Prompt 1: <|image|><|begin_of_text|>Update a JSON representation based on a graphical difference relationship.
You are given:
1.) A JSON object containing the agent's current state.
2.) An image illustrating the relationship between the current and future quantity. Both quantities are represented as '?'.
Based on this information, you should:
A.) Extract the full JSON object.
B.) Infer the future quantity using the information from the image.
C.) Return the updated JSON object, ensuring that the 'quantity' field contains the final numerical value, not an equation or expression.
This is the JSON object: '{"agent": "Charlene", "quantity": 13, "entity": "lamps", "attribute": "", "unit": ""}'

Output Step 1: The image shows that the quantity is reduced by 2. Therefore, the future quantity is 13 - 2 = 11. The updated JSON object is: '{"agent": "Charlene", "quantity": 11, "entity": "lamps", "attribute": "", "unit": ""}'

Update Prompt 2: <|image|><|begin_of_text|>Update a JSON representation based on a graphical difference relationship.
You are given:
1.) A JSON object containing the agent's current state.
2.) An image illustrating the relationship between the current and future quantity. Both quantities are represented as '?'.
Based on this information, you should:
A.) Extract the full JSON object.
B.) Infer the future quantity using the information from the image.
C.) Return the updated JSON object, ensuring that the 'quantity' field contains the final numerical value, not an equation or expression.
This is the JSON object: '{"agent": "Charlene", "quantity": 11, "entity": "lamps", "attribute": "", "unit": ""}'

Output Step 2: The image shows that the quantity is increased by 6. Therefore, the future quantity is 11 + 6 = 17. 
The updated JSON object is: '{"agent": "Charlene", "quantity": 17, "entity": "lamps", "attribute": "", "unit": ""}'
\end{minted}
\end{small}
\caption{Llama3.2-11B output for Vision Mental Modeling ($\VMM$).}
\label{fig:prompt-VMM}
\end{listing*}

\begin{listing*}

\begin{small}
\captionsetup{justification=centering, labelfont=bf, font=small}
\begin{minted}[fontsize=\footnotesize, frame=lines, framesep=2mm, baselinestretch=1.2, breaklines, breaksymbolleft={}, breaksymbolright={}]{text}
Problem: Charlene has 13 lamps. Then Ingaborg got 2 more lamps from Charlene. Modesta gave Charlene 6 lamps. How many lamps does Charlene have totaled now?

Initial Prompt: Extract structured information from the following sentence and update an empty JSON object with the categories: 'agent', 'quantity', 'entity', 'attribute', and 'unit'. If a category is not applicable, leave it as an empty string. Do not introduce additional categories.
This is the sentence: 'Charlene has 13 lamps.'.

Initial Output (Full python code generated to be found in Listing 4): Here is the output of the code:
```
{
  "agent": "Charlene",
  "quantity": "13",
  "entity": "lamps",
  "attribute": "",
  "unit": ""
}
```

Update Prompt 1: Update the given JSON representation based on a described quantity change.
You are given:
1.) A JSON object representing the agent's current state.
2.) A natural language sentence that describes a change in quantity for the agent.
Based on this information, you should:
A.) Extract the full JSON object.
B.) Interpret the sentence to determine how the quantity changes.
C.) Return the updated JSON object, ensuring that the 'quantity' field contains the final numerical value, not an equation or expression.
This is the natural language sentence: 'Then Ingaborg got 2 more lamps from Charlene.'.
This is the JSON object: '{"agent": "Charlene", "quantity": "13", "entity": "lamps", "attribute": "", "unit": ""}'.

Output Step 1: The final answer is: 
{"agent": "Charlene", "quantity": "15", "entity": "lamps", "attribute": "", "unit": ""} 
Note: The final answer is a JSON object.  Please do not forget to add the curly brackets.

Update Prompt 2: Update the given JSON representation based on a described quantity change.
You are given:
1.) A JSON object representing the agent's current state.
2.) A natural language sentence that describes a change in quantity for the agent.
Based on this information, you should:
A.) Extract the full JSON object.
B.) Interpret the sentence to determine how the quantity changes.
C.) Return the updated JSON object, ensuring that the 'quantity' field contains the final numerical value, not an equation or expression.
This is the natural language sentence: 'Modesta gave Charlene 6 lamps.'.
This is the JSON object: '{"agent": "Charlene", "quantity": "15", "entity": "lamps", "attribute": "", "unit": ""}'.

Output Step 2: The final answer is: 
{"agent": "Charlene", "quantity": "21", "entity": "lamps", "attribute": "", "unit": ""} 
Explanation: 
The sentence 'Modesta gave Charlene 6 lamps.' implies that Charlene received 6 lamps from Modesta. To update the quantity, we need to add 6 to the current quantity of Charlene's lamps. The current quantity is 15, so the new quantity will be 15 + 6 = 21. The updated JSON object will be '{"agent": "Charlene", "quantity": "21", "entity": "lamps", "attribute": "", "unit": ""}'. 
Note: The unit of the quantity is not specified in the sentence, so it remains empty. 
\end{minted}
\end{small}
\caption{Llama3.2-11B output for Text Mental Modeling ($\TMM$).}
\label{fig:prompt-TMM}
\end{listing*}

\begin{listing*}
\begin{small}
\captionsetup{justification=centering, labelfont=bf, font=small}
\begin{minted}[fontsize=\footnotesize, frame=lines, framesep=2mm, baselinestretch=1.2, breaklines, breaksymbolleft={}, breaksymbolright={}]{text}
Here is the JSON object to be updated:
```
{
  "agent": "",
  "quantity": "",
  "entity": "",
  "attribute": "",
  "unit": ""
}
```
Here is the Python code to solve the problem:
```python
import re
def extract_info(sentence):
    # Initialize the JSON object
    info = {
        "agent": "",
        "quantity": "",
        "entity": "",
        "attribute": "",
        "unit": ""
    }
    # Extract the agent
    match = re.search(r'\b\w+\b', sentence)
    if match:
        info['agent'] = match.group()
    # Extract the quantity
    match = re.search(r'\d+', sentence)
    if match:
        info['quantity'] = match.group()
    # Extract the entity
    match = re.search(r'\b\w+\b', sentence)
    if match:
        info['entity'] = match.group()
    # Extract the attribute
    match = re.search(r'\b\w+\b', sentence)
    if match:
        info['attribute'] = match.group()
    # Extract the unit
    match = re.search(r'\b\w+\b', sentence)
    if match:
        info['unit'] = match.group()
    return info
# Test the function
sentence = 'Charlene has 13 lamps.'
print(extract_info(sentence))
```
Here is the output of the code:
```
{
  "agent": "Charlene",
  "quantity": "13",
  "entity": "lamps",
  "attribute": "",
  "unit": ""
}
```
\end{minted}
\end{small}
\caption{Full initial output generated by Llama3.2-11B during $\TMM$.}
\label{lst:python}
\end{listing*}

\begin{listing*}

\begin{small}
\captionsetup{justification=centering, labelfont=bf, font=small}
\begin{minted}[fontsize=\footnotesize, frame=lines, framesep=2mm, baselinestretch=1.2, breaklines, breaksymbolleft={}, breaksymbolright={}]{text}
Problem: Charlene has 13 lamps. Then Ingaborg got 2 more lamps from Charlene. Modesta gave Charlene 6 lamps. How many lamps does Charlene have totaled now?

Vision-only Inference Prompt: <|image|><|begin_of_text|>You are tasked with interpreting a graphical representation of a difference relationship. 
To do so, answer the question below. You are not given the answer explicitly, but you should infer it from the graphical representation.
Based on this information, answer the following question.
Q: How many lamps does Charlene have totaled now?
A: 

Output for Llama3.2-11B: 19
**Step 1: Identify the initial quantity of lamps Charlene has.**
*   The initial quantity of lamps Charlene has is 13.
**Step 2: Calculate the change in the quantity of lamps.**
*   The change in the quantity of lamps is -2.
**Step 3: Calculate the new quantity of lamps Charlene has.**
*   The new quantity of lamps Charlene has is 13 - 2 = 11.
**Step 4: Calculate the change in the quantity of lamps again.**
*   The change in the quantity of lamps is +6.
**Step 5: Calculate the final quantity of lamps Charlene has.**
*   The final quantity of lamps Charlene has is 11 + 6 = 17.
**Step 6: Calculate the change in the quantity of lamps again.**
*   The change in the quantity of lamps is -2.
**Step 7: Calculate the final final quantity of lamps Charlene has.**
*   The final final quantity of lamps Charlene has is 17 - 2 = 15.
**Step 8: Calculate the change in the quantity of lamps again.**
*   The change in the quantity of lamps is +6.
**Step 9: Calculate the final final final quantity of lamps Charlene has.**
*   The final final final quantity of lamps Charlene has is 15 + 6 = 21.
**Step 10: Calculate the change in the quantity of lamps again.**
*   The change in the quantity of lamps is -2.
**Step 11: Calculate the final final final final quantity of lamps Charlene has.**
*   The final final final final quantity of lamps Charlene has is 21 - 2 = 19.

Output for Llama3.2-90B: 13 + 6 = 19. Charlene has 19 lamps. The -2 lamps are not included in the total because they are not owned by Charlene. They are owned by an unknown entity. The +6 lamps are included in the total because they are owned by Charlene. The 13 lamps are included in the total because they are owned by Charlene. The unknown entity's lamps are not included in the total because they are not owned by Charlene. Therefore, the total number of lamps owned by Charlene is 19. The answer is 19.
\end{minted}
\end{small}
\caption{Llama3.2-11B and Llama3.2-90B output for Vision-only Inference (\texttt{VoI}).}
\label{fig:prompt-VoI}
\end{listing*}

\begin{listing*}

\begin{small}
\captionsetup{justification=centering, labelfont=bf, font=small}
\begin{minted}[fontsize=\footnotesize, frame=lines, framesep=2mm, baselinestretch=1.2, breaklines, breaksymbolleft={}, breaksymbolright={}]{text}
Problem: Charlene has 13 lamps. Then Ingaborg got 2 more lamps from Charlene. Modesta gave Charlene 6 lamps. How many lamps does Charlene have totaled now?

Chain-of-Thought Prompt: Charlene has 13 lamps. Then Ingaborg got 2 more lamps from Charlene. Modesta gave Charlene 6 lamps. How many lamps does Charlene have totaled now?

Output for Llama3.2-11B: 13 + 2 + 6 = ?
## Step 1: Calculate the initial number of lamps Charlene has.
Charlene starts with 13 lamps.   
## Step 2: Calculate the number of lamps Charlene has after Ingaborg got 2 more lamps from her.
Charlene now has 13 + 2 = 15 lamps.    
## Step 3: Calculate the total number of lamps Charlene has after Modesta gave her 6 lamps.
Charlene now has 15 + 6 = 21 lamps. 
The final answer is: $\boxed{21}$

Output for Llama3.2-90B: Charlene had 13 lamps. She gave 2 to Ingaborg. So she had 13 - 2 = 11 lamps. Then Modesta gave her 6 lamps. So Charlene has 11 + 6 = 17 lamps. The answer is 17.
## Step 1: Calculate the number of lamps Charlene had after giving 2 lamps to Ingaborg.
Charlene had 13 lamps and gave 2 to Ingaborg, so she had 13 - 2 = 11 lamps. 
## Step 2: Calculate the total number of lamps Charlene has after Modesta gave her 6 lamps.
Charlene had 11 lamps and Modesta gave her 6 lamps, so Charlene now has 11 + 6 = 17 lamps.
The final answer is: $\boxed{17}$
\end{minted}
\end{small}
\caption{Llama3.2-11B and Llama3.2-90B output for Chain-of-Thought (\texttt{CoT}).}
\label{fig:prompt-cot}
\end{listing*}

\end{document}